\documentclass[12pt]{article}

\usepackage[margin=1in]{geometry}
\usepackage{graphicx}
\usepackage{cite}
\usepackage{amsmath}
\usepackage{hyperref} 
\usepackage{xcolor}
\usepackage{epsfig}
\usepackage{comment}
\usepackage{booktabs}
\usepackage{adjustbox}
\usepackage{threeparttable}
\usepackage{colortbl}
\usepackage{array}
\usepackage{caption}
\usepackage{float}
\usepackage{tikz}
\usepackage{pgfplots}
\pgfplotsset{compat=1.17}
\usepackage{filecontents}
\usepackage{setspace}
\usepackage{titlesec}
\usepackage{titling} 
\usepackage{booktabs}
\usepackage{caption}
\usepackage{tikz}
\usetikzlibrary{tikzmark}

\titleformat{\section}{\large\bfseries}{\thesection.}{0.5em}{}
\titleformat{\subsection}{\normalsize\bfseries}{\thesubsection}{0.5em}{}

\pretitle{\begin{center}\Large\bfseries}
\posttitle{\par\end{center}}
\preauthor{\begin{center}}
\postauthor{\par\end{center}}
\predate{\begin{center}}
\postdate{\par\end{center}}

\title{\textbf{Evaluating an Adaptive Multispectral Turret System for Autonomous Tracking Across Variable Illumination Conditions}}

\author{
Aahan Sachdeva\textsuperscript{1}\thanks{Indicates major responsibility}, 
Dhanvinkumar Ganeshkumar\textsuperscript{1}\footnotemark[1], 
James E. Gallagher\textsuperscript{1},\\
Tyler Treat\textsuperscript{1},
Edward J. Oughton\textsuperscript{1}\\[0.5em]
\textsuperscript{1}Department of Geography \& Geoinformation Science,\\
George Mason University, Fairfax, VA, USA\\
Corresponding author: James E. Gallagher (\url{jgalla5@gmu.edu})\\
}

\date{December 2025}

\begin{document}

\maketitle

\begin{abstract}
Autonomous robotic platforms are playing a growing role across the emergency services sector, supporting missions such as search and rescue operations in disaster zones and reconnaissance. However, traditional red-green-blue (RGB) detection pipelines struggle in low-light environments, and thermal-based systems lack color and texture information. To overcome these limitations, we present an adaptive framework that fuses RGB and long-wave infrared (LWIR) video streams at multiple fusion ratios and dynamically selects the optimal detection model for each illumination condition. We trained 33 You Only Look Once (YOLO) models on over 22,000 annotated images spanning three light levels: no-light (\textless 10 lux), dim-light (10–1000 lux), and full-light (\textgreater 1000 lux). To integrate both modalities, fusion was performed by blending aligned RGB and LWIR frames at eleven ratios, from full RGB (100/0) to full LWIR (0/100) in 10\% increments. Evaluation showed that the best full-light model (80/20 RGB--LWIR) and dim-light model (90/10 fusion) achieved 92.8\% and 92.0\% mean confidence; both significantly outperformed the YOLOv5 nano (YOLOv5n) and YOLOv11 nano (YOLOv11n) baselines. Under no-light conditions, the top 40/60 fusion reached 71.0\%, exceeding baselines though not statistically significant. Adaptive RGB--LWIR fusion improved detection confidence and reliability across all illumination conditions, enhancing autonomous robotic vision performance.
\end{abstract}


\section{Introduction}

Autonomous vision systems are increasingly vital for critical operations such as search and rescue, disaster response, and defense surveillance \cite{ref61,sampedro2019,ref69}. These applications demand reliable, real-time object detection across a wide range of environmental conditions, where lighting, weather, and visibility can change unpredictably. Conventional computer vision pipelines based solely on visible-spectrum RGB imagery often suffer performance degradation in low-light conditions \cite{ref3}. Conversely, long-wave infrared (LWIR) thermal imaging remains accurate in darkness but lacks the texture and color information that RGB sensors provide \cite{ref64}.

\begin{figure}[H]
    \vspace{-60pt}
    \centering
    \includegraphics[width=0.85\linewidth]{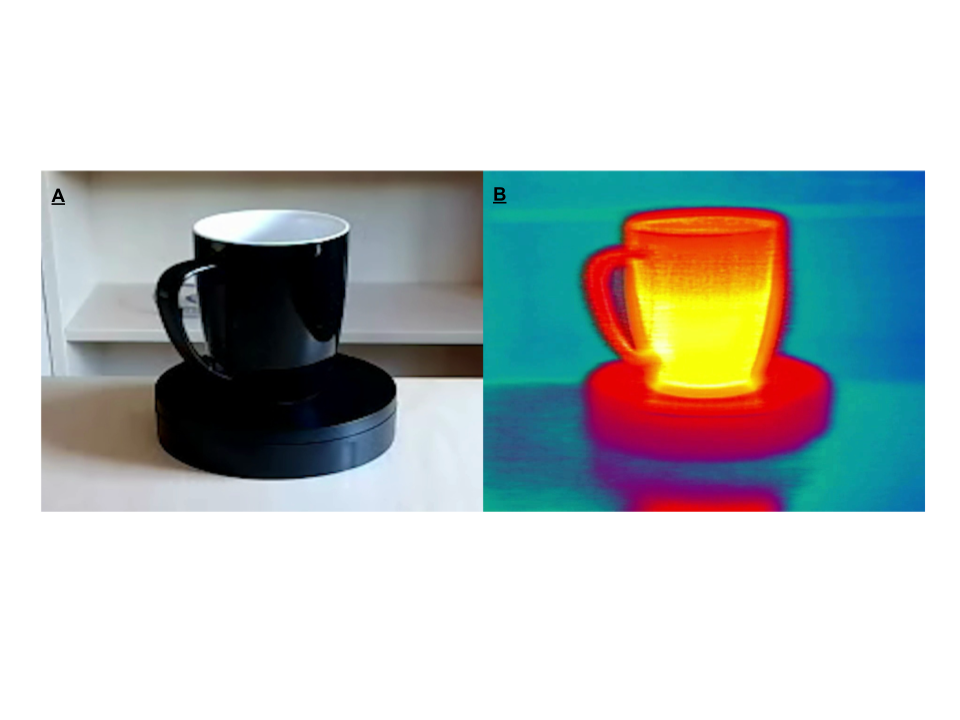}
    \vspace{-60pt}
    \caption{Example images illustrating the two modalities: RGB image (A) and corresponding LWIR thermal image (B) of the same scene.}
    \vspace{-6pt}
    \label{fig:fusion_example}
\end{figure}

The complementary strengths of RGB and LWIR sensing have driven growing interest in multimodal perception systems \cite{ref5_new,ref6_new}. Figure~\ref{fig:fusion_example} illustrates an example of these two modalities, showing an RGB image (A) on the left alongside its corresponding LWIR thermal image (B) of the same scene. However, achieving effective fusion in real-world environments presents several challenges:  
(1) fusion strategies must preserve discriminative features from both modalities without introducing distortions; 
(2) computational efficiency is critical for deployment on embedded or resource-constrained platforms; and 
(3) adaptability is required to handle dynamic illumination changes without human intervention. 
Existing approaches often address only a subset of these requirements, limiting their accuracy in operational environments.

Deep learning–based object detection, particularly models in the You Only Look Once (YOLO) family, have demonstrated strong performance in single-modality vision tasks \cite{ref67,ref68}. Extending such architectures to handle multimodal fusion while maintaining real-time capability offers a promising pathway toward reliable autonomous vision. This integration requires careful consideration of fusion levels, feature alignment, and model selection strategies that adapt to environmental context.

In this work, we make three primary contributions. First, we develop a practical RGB--LWIR fusion approach that blends aligned images at varying opacity levels, improving target visibility across diverse lighting conditions. Second, we fine-tune and evaluate 33 YOLO models to identify the top-performing fusion configurations for each lighting condition, trained on over 22,000 mug-illumination images, which we open-source. Third, we implement an adaptive detection framework that selects among lighting-specific models based on ambient illumination.

Through these contributions, we investigate how RGB--LWIR fusion ratio, illumination level, and model specialization all affect detection confidence in a real-time turret system.

The remainder of this paper is organized in the following sections. Section~\ref{sec:literature_review} reviews related work in multimodal perception and adaptive object detection, while section~\ref{sec:methodology} details the training and system architecture. Section~\ref{sec:results} presents quantitative and qualitative performance evaluations. Section~\ref{sec:discussion} discusses limitations and implications. Finally, section~\ref{sec:conclusion} concludes the paper and outlines future research directions.

\section{Literature Review}
\label{sec:literature_review}

\subsection{Multispectral Object Detection and Sensor Fusion}

Multispectral object detection builds on a broader body of work in multi-sensor perception and fusion for autonomous systems. 

Feng et al.\ provided a foundational survey of deep multi-modal object detection and semantic segmentation for autonomous driving, covering typical sensor suites including camera, Light Detection and Ranging (LiDAR), radar, Global Navigation Satellite System (GNSS), and Inertial Measurement Unit (IMU). They also organized fusion strategies by what to fuse, how to fuse it, and when fusion should occur in the pipeline~\cite{ref4}. Other surveys focus specifically on adverse-weather perception~\cite{ref5}, camera--LiDAR fusion~\cite{ref6}, and multi-sensor fusion plus cooperative perception~\cite{ref7,ref12}. Together, these works show that combining complementary modalities generally improves accuracy to occlusions and poor visibility. However, it also introduces practical challenges around sensor calibration and real-time computation. More recently, research in geoscience has emphasized the importance of integrating multiple data sources and designing systems that can adapt to different sensing conditions, particularly in remote sensing and Earth-observation applications~\cite{ref76}.

Gallagher and Oughton extended this perspective into the YOLO ecosystem with a large-scale survey of multispectral YOLO detectors from 2020--2024~\cite{ref1}. Screening 400 studies, they report that RGB--infrared combinations, especially RGB--LWIR, dominate current work. They also found YOLOv5 and its variants remain the most common backbones for multispectral modification, and that ground-based data still outweigh aerial and satellite platforms~\cite{ref1}. Their analysis also surfaces several gaps that are directly relevant to our research: a mismatch between mean average precision (mAP) on curated benchmarks and performance in cluttered, real-world scenes, limited exploration of multi-channel 16-bit grayscale inputs, and relatively few studies that quantify latency–accuracy trade-offs for real-time deployment. 

Beyond detection-specific surveys, Senel et al.\ proposed a modular fusion framework for real-time multi-object tracking that operates at the object-list level across cameras, LiDAR, and radar~\cite{ref13}. Their results confirm that late, object-level fusion can be computationally efficient and robust to individual sensor failures, complementing feature-level fusion approaches and informing how multispectral detectors can feed downstream tracking in a turret or autonomous vehicle (AV) stack.

\subsection{RGB--Thermal Fusion Architectures}

Within multispectral perception, RGB--thermal (RGB--T) fusion emerged as a dominant configuration for 24/7 detection in low-light or adverse conditions. Early work on multispectral pedestrian detection systematically compared fusion stages in Faster Region-based Convolutional Neural Network (Faster R-CNN) detectors and showed that simple input-level fusion often performed worse than more carefully designed halfway or score-level fusion~\cite{ref9}. Building on this, illumination-aware architectures such as Illumination-Aware Faster R-CNN used a dedicated illumination estimation network and learned gating that weighted RGB versus thermal branches depending on scene brightness, and improved performance on the multispectral pedestrian benchmark developed by the Korea Advanced Institute of Science and Technology (KAIST), particularly at night~\cite{ref20,ref24}.

A large body of work then focused specifically on RGB--T detectors that operated in real time. The Multi-modal Attention Fusion YOLO (MAF-YOLO) model compressed dual Darknet-53 backbones and introduced illumination-aware feature weighting plus dual attention for multispectral pedestrian detection, reaching near two-stage accuracy at single-stage speed~\cite{ref11}. Multispectral Object Detection Algorithm for Complex Road Scenes (GMD-YOLO) augmented a dual-channel Cross Stage Partial Darknet (CSPDarknet) with multilevel feature fusion and dual feature modulation to better handle small and occluded targets~\cite{ref23}, while Multimodal Multi-scale YOLO Fusion Network (MMYFNet) integrated cosine-similarity-based feature separation and refinement into a YOLOv8-style dual-branch design for RGB--infrared (IR) remote-sensing detection~\cite{ref25}. Multimodal Representation Distillation–You Only Look Once (MRD-YOLO), Two-Stream Fusion–You Only Look Once (TF-YOLO), Feature-Map Pyramid Fusion Network (FMPFNet), Self-Adaptive Multispectral–You Only Look Once (SAMS-YOLO), You Only Look Once with Cross-Channel Attention Module (YOLO-CCAM), You Only Look Once version 11 for Red–Green–Blue–Thermal imagery (YOLOv11-RGBT), Progressive Object Network (PONet), and Feature-Quality-Driven Network (FQDNet) each represented different design choices in this space, incorporating frequency-domain fusion, transformer-based cross-modal attention, pixel-level luminance-weighted fusion, group-shuffled multi-receptive attention, cross-modality context attention, and flexible mid-fusion strategies respectively~\cite{ref26,ref27,ref28,ref29,ref34,ref35, ref72,ref73}. Taken together, these methods showed that mid-level fusion at selected feature scales often provided a better trade-off than purely early or late fusion, that attention and gating modules were important for handling changes in modality quality across illumination and weather, and that modifications to standard YOLO backbones could still yield strong multispectral gains without large computational cost \cite{ref74}.

A separate line of work addressed pixel-level infrared--visible fusion for human viewing and downstream perception. Parallel Interactive Attention Fusion (PIAFusion) used an illumination-aware loss and progressive cross-modality differential fusion modules to generate fused infrared--visible images that preserved both thermal features and visible textures~\cite{ref10}. Xu et al.\ proposed a generative adversarial network (GAN) + residual network (ResNet)-based fusion network that encouraged fused outputs to inherit fine visible textures while maintaining infrared intensity, and it performed better than classical pyramid and wavelet-based fusion under multiple objective metrics~\cite{ref19}. Alldieck et al.\ instead performed context-aware fusion at the segmentation level for traffic monitoring, computing per-frame modality weights from sun elevation, shadow likelihood, thermal entropy, and foreground stability to fuse soft foreground maps from RGB and thermal background subtractors~\cite{ref16}. Serrano-Cuerda et al.\ moved further toward sensor selection, using simple confidence measures on visible and thermal imagery to decide when to rely on each camera rather than fusing them, and achieved strong outdoor detection of humans with relatively low computational cost~\cite{ref17}. These studies showed that adaptive, context-aware weighting often worked better than simple averaging, and similar ideas appeared in many RGB--T detection architectures.

\subsection{Datasets, Labeling Pipelines, and Benchmarks}

Multispectral detection depends heavily on labeled datasets and efficient annotation workflows. Gallagher and Oughton developed an open RGB, LWIR, and fused RGB--LWIR unmanned aerial vehicle (UAV) dataset for vehicle and person detection. They co-aligned the sensors and varied altitude and time of day to identify the conditions under which fusion is most effective~\cite{ref2}. Their dataset provides a controlled way to compare RGB-only, LWIR-only, and fused detection. Speth et al.\ collected aerial RGB and far-infrared (FIR) imagery onboard drones for scene understanding and person detection, releasing the Okutama Drone, Swiss Drone, and Normalized Intensity Information – Cross-Unit (NII-CU) multispectral person datasets. They show that a simple four-channel early-fusion YOLOv3 can outperform late-fusion ensembles for onboard detection~\cite{ref14}. Building on the broader need for UAV-based thermal–visible datasets, Zhang et al.\ introduce the RGBT-3M dataset, which contains over 17k RGB–thermal pairs for smoke, fire, and person detection. They propose a mid-fusion YOLOv11 variant with cross-modal interaction and patch-aware splicing modules that improve no-light and occlusion robustness~\cite{ref18}.

Beyond RGB–T datasets, Gani et al.\ constructed a small RGB–near-infrared (NIR)–thermal collection to evaluate YOLOv3 under different spectral combinations, showing that sensor misalignment and limited sample size can offset the expected gains from adding more modalities~\cite{ref33}. In a different domain, Qingqing et al.\ released a heterogeneous multi-LiDAR dataset containing both spinning and solid-state units alongside RGB–depth data, enabling cross-sensor benchmarking for simultaneous localization and mapping (SLAM) and odometry~\cite{ref21}. For camouflaged object detection and saliency, Li et al.\ introduced Multispectral Camouflaged Object Detection (MCOD), an eight-channel multispectral benchmark featuring tiny objects and challenging backgrounds~\cite{ref39}, while Wang et al.\ released UVT20K, a large-scale, alignment-free RGB–T saliency dataset with 20k pairs and detailed attribute annotations~\cite{ref31}.

Labeling efficiency is a major bottleneck for multispectral datasets because each modality must be spatially aligned and annotated. Gallagher et al.\ addressed this with the Multispectral Automated Transfer Technique (MATT), which uses the Segment Anything Model to segment objects in RGB frames, transfers masks to paired LWIR and fused frames, and then trains YOLO models with minor human verification, reducing labeling time substantially with only modest mAP loss~\cite{ref3}. Recent surveys of the Segment Anything Model (SAM) further highlight its effectiveness as a general-purpose segmentation tool, while noting the importance of domain-specific adaptations when applying it to specialized settings such as remote sensing and multispectral imagery~\cite{ref77}. More recently, Self-Adaptive Multispectral Object Detection (SAMSOD) and related SAM-based pipelines revisit how to fine-tune SAM-like models for RGB--T saliency, introducing unimodal supervision and gradient deconfliction techniques that stabilize training and improve boundary precision~\cite{ref38, ref78}. Taken together, these works highlight how automated or semi-automated label transfer across modalities is becoming essential for scaling multispectral datasets, directly motivating the efficient data pipelines required for practical systems like our turret.

MSOD-Bench offers a complementary benchmarking perspective by re-implementing and standardizing more than twenty RGB–T detection models, including many of the YOLO-based variants discussed above across datasets such as KAIST Multispectral Pedestrian Dataset (KAIST), the Low-Light Visible–Infrared Person Dataset (LLVIP), the DroneVehicle, and the FLIR Thermal Dataset for Advanced Driver Assistance Systems (FLIR ADAS)~\cite{ref30}. It also consolidates a practical “bag of tricks,” ranging from methods such as re-clustering to cosine-annealed learning rates and balanced augmentation, which help close performance gaps without modifying the underlying architectures. This benchmark is particularly important to the research reported in this paper, because it shows that a significant portion of the performance variation among multispectral detectors arises from training procedures and data handling, rather than network pipeline alone.

\subsection{Robustness to Illumination, Weather, and Modality Imbalance}

A core motivation for multispectral perception is robustness to changing illumination, weather, sensor health, and adversarial or data-scarce conditions \cite{ref70, ref71}. Zhang et al.\ surveyed sensor degradation and countermeasures under rain, fog, and snow across cameras, LiDAR, radar, and Global Navigation Satellite System / inertial navigation system (GNSS/INS), highlighting the need for fusion strategies that explicitly account for modality-specific failure modes~\cite{ref5}. Many RGB--T detectors in the literature addressed this by introducing illumination-aware or context-aware gating modules that make thermal features more important at night and RGB features more important in full-light, as in illumination-aware Faster R-CNN, MAF-YOLO, PIAFusion, and context-aware traffic monitoring~\cite{ref17,ref20,ref24}. Other models added transformer-based fusion blocks and cross-modality context attention to better exploit long-range dependencies and handle occlusions and low visibility~\cite{ref23,ref27,ref29}.

Recent work went further by focusing on domain adaptation and modality imbalance. D3T introduced a dual-teacher mean-teacher framework for unsupervised domain adaptation from RGB to thermal detection, alternating training between visible and infrared domains to gradually reduce the gap between the two~\cite{ref36}. Tian et al.\ designed a base--auxiliary detector architecture for extreme modality imbalance, where one modality could be heavily degraded or entirely missing, and combined pseudo-degradation, exponential moving average (EMA)-based supervision, and a quality-aware modality interaction module to maintain performance under sensor failures~\cite{ref32}. RDCRNet addressed cross-modal distribution gaps and misalignment by explicitly remapping and refining modality-specific statistics and features before fusion, and showed clear improvements on misaligned RGB--T datasets~\cite{ref22}. Multi-stage frameworks such as MO R-CNN and MMYFNet incorporated deformable convolutions, alignment modules, and label-fusion strategies to handle heterogeneous sensors and inconsistent annotations in remote-sensing imagery~\cite{ref25,ref37}.

Finally, several RGB--T saliency models, such as MCNet, PCNet, and SAMSOD, were developed to address alignment error and gradient conflicts. These models used mirror-complementary supervision, progressive correlation modules, and decoupled adapters to balance RGB and thermal contributions more evenly~\cite{ref31,ref38,ref40}. Although these studies focused on saliency rather than detection, their design choices alignment-aware modules, explicit modality balancing, and training strategies that prevent the RGB branch from dominating remain useful when building detectors for real-world deployments.

\subsection{Aerial, Road, and Turret-Like Multispectral Applications}

Application domains for RGB–T detection span aerial monitoring, autonomous driving, surveillance, and inspection tasks such as anomaly detection in energy infrastructure \cite{ref75}. In aerial contexts, Gallagher and Oughton showed through drone-mounted RGB–LWIR experiments that early fusion substantially improved pre-sunrise and post-sunset detection and remained competitive in full-light, whereas LWIR-only performance degraded rapidly with altitude~\cite{ref2}. Speth et al.\ presented an onboard pipeline that combined RGB–FIR segmentation for landing-site assessment with YOLO-based person detection, and quantified accuracy–latency trade-offs on embedded GPUs~\cite{ref14}. Zhang et al.’s wildfire dataset and CP-YOLOv11-MF model demonstrated that mid-fusion and cross-modal interaction enhanced detection of small or partially occluded fire and smoke signatures~\cite{ref18}. Building on this line of work, FMPFNet adapted lightweight luminance-weighted pixel-level fusion in the luminance--chrominance (YUV) space and used dynamic modality dropout to reduce over-reliance on RGB in low-light conditions~\cite{ref28}.

In road environments, a progression of YOLO-based RGB–T detectors, MAF-YOLO, GMD-YOLO, MRD-YOLO, TF-YOLO, YOLO-CCAM, SAMS-YOLO, MMYFNet, and YOLOv11-RGBT, along with RadNet-style multimodal frameworks showed that mid-/back-end fusion, cross-modal attention, and illumination-aware weighting were central to robust pedestrian, vehicle, and small-object detection under full-light, no-light, and adverse-weather conditions~\cite{ref4,ref12,ref25}. Additional work on multi-scale fusion, contrast-adaptive feature mixing, and modality-selection modules further refined these capabilities~\cite{ref26,ref30}. These systems were commonly evaluated on KAIST, LLVIP, DroneVehicle, FLIR ADAS, M3FD, and UTokyo, reflecting their focus on automotive, on-vehicle deployment.

Our work is closest in spirit to these Red, Green, Blue – Thermal You Only Look Once (RGB–T YOLO) detectors and aerial studies but targets a distinct operational regime: a fixed pan–tilt turret that must track and engage small objects (coffee cups) at close range across full-light, dim-light, and no-light conditions. Whereas prior work typically evaluated a single fusion strategy or architecture per dataset, we systematically vary RGB–LWIR fusion ratios on YOLOv11 backbones, measure detection and tracking performance across illumination regimes, and tie these results to a real-time actuation task. The literature reviewed above showed the importance of adaptive, illumination-aware fusion, robust training strategies, and curated datasets; our experiments extend these insights to a turret platform and quantify how simple linear RGB–LWIR pixel-level fusion interacts with modern YOLO architectures under realistic hardware and latency constraints.

\section{Methodology}
\label{sec:methodology}

\subsection{System hardware overview}
\label{sec:system_overview}
The experimental platform is an autonomous turret prototype built around a Raspberry Pi~5. Two stepper motors provide pan and tilt motion. Each motor is driven by a DRV8825 stepper driver mounted on a stepper motor HAT connected to the Raspberry Pi.

Perception relies on two imaging sensors and one illumination sensor. A visible-spectrum RGB camera (Emeet S600 webcam) captures color and texture information, and a long-wave infrared thermal camera (FLIR Vue Pro R) captures thermal signatures. An ambient light sensor (Adafruit VEML7700) measures scene illumination in lux. The Raspberry Pi captures the sensor data and forwards it to a remote server via a FastAPI endpoint. The server performs the computationally intensive object-detection inference and returns the detection results to the Pi, which converts the detected target positions into pan and tilt commands for the stepper motors.

\begin{figure}[H]
    \centering
    \includegraphics[width=\textwidth,keepaspectratio]{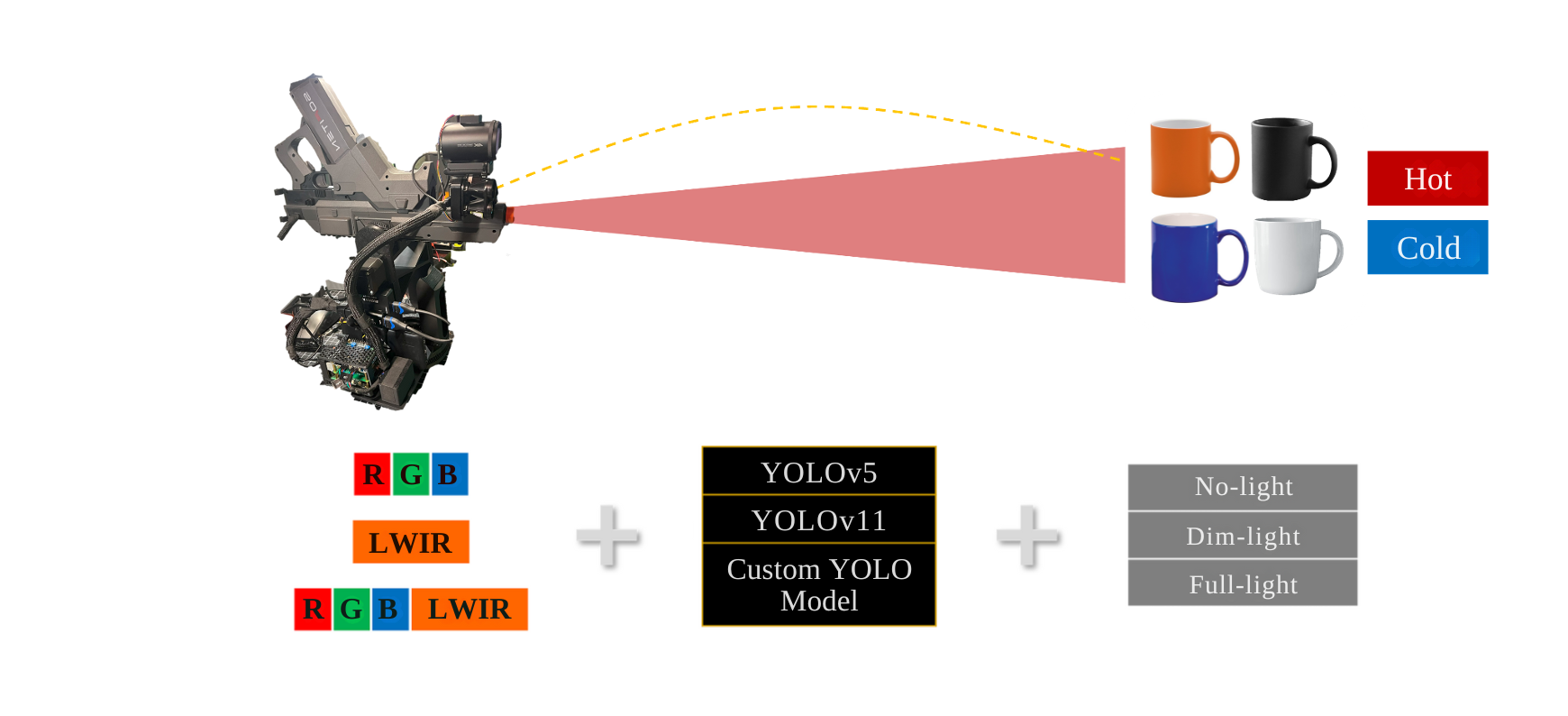}
    \caption{Hardware and system overview of the autonomous turret, illustrating the four independent experimental variables: fusion level, illumination category, model architecture, and target characteristics}
    \label{fig:turret_system}
\end{figure}

Figure~\ref{fig:turret_system} provides a visual summary of the full turret system and highlights the four primary experimental variables: fusion level, illumination category, object detection model architecture, and target characteristics.

\subsection{Experimental variables}
\label{sec:experimental_variables}

We control four primary variables in the study: fusion level, illumination category, model architecture, and target characteristics. 

Fusion level varies by blending the RGB and thermal images. For each blending weight $\alpha \in \{1.0, 0.9, \ldots, 0.0\}$, we compute a fused frame as defined in Equation~\ref{eq:fusion}:

\begin{equation}
\label{eq:fusion}
F = \alpha \cdot \mathrm{RGB}_{\text{norm}} + (1-\alpha) \cdot \mathrm{LWIR}_{\text{norm}} .
\end{equation}

Here, $\alpha = 1.0$ corresponds to a purely RGB input and $\alpha = 0.0$ corresponds to a purely thermal input. $\mathrm{RGB}_{\text{norm}}$ and $\mathrm{LWIR}_{\text{norm}}$ represents the standard RGB and LWIR images for the same frame, which is blended by adjusting the opacity of each image to control its influence on the final fused image. This produces eleven distinct fusion levels spanning from RGB-only to LWIR-only.

LWIR frames are recorded and processed using the thermal camera's default color palette, and this palette is held constant across all fusion levels and lighting conditions. 

Illumination category is determined from lux readings which the VEML7700 sensor provides. Based on these measurements, we divide the scenes into three ranges: full-light, defined as lux $> 1000$; dim-light, defined as $10 \leq \text{lux} \leq 1000$; and no-light, defined as lux $< 10$. These thresholds are chosen to roughly separate bright full-light conditions, indoor or twilight conditions, and near-complete darkness.

For the model architecture variable, we fine-tune thirty-three YOLOv11 nano detectors, training one model for each combination of fusion level and illumination category. In addition to these task-specific models, we include two pre-trained detectors as baselines: a YOLOv11 nano model and a YOLOv5 nano model, which are both trained on the Microsoft Common Objects in Context (COCO) dataset. These baselines provide a reference for how a generic RGB-only detector performs relative to our specialized multispectral models.

Finally, target characteristics are varied through mug color. Four mug colors (white, black, orange, and dark blue) are used during training, while two additional colors (teal and yellow) are reserved exclusively for out-of-distribution testing. This design allows us to measure how well each model generalized to previously unseen target appearances.

\subsection{Data collection}
\label{sec:data_collection}
Synchronized RGB and thermal video streams are recorded while presenting target mugs on a rotating platform. Each rotation captures a full 360-degree view over eight seconds, balancing smooth multi-angle coverage with minimal motion blur and practical acquisition time. Dynamic following trials last ten seconds, providing sufficient temporal data to assess detection stability and turret control behavior.

The object-to-camera distance is held at approximately three feet (0.9 m) for all recordings, ensuring consistent target scale across trials. Approximately 22,000 fused images are produced from the recordings. All images are annotated with bounding boxes for the single class, ``mug,'' using LabelImg and the Computer Vision Annotation Tool (CVAT). The dataset is split into training and validation subsets with a 75\% / 25\% ratio.

\subsection{Image alignment and fusion}
\label{sec:image_fusion}
Accurate alignment between RGB and thermal frames is critical for reliable fusion. Frames are synchronized temporally and aligned spatially; The fusion procedure followed the linear blending formula in Section~\ref{sec:experimental_variables}. Figure~\ref{fig:fusion_levels} illustrates example frames across the fusion spectrum.

\begin{figure}[H]
    \centering
    \includegraphics[width=\textwidth,keepaspectratio]{fusion_levels.png}
    \caption{Representative illustration of the eleven fusion levels used for training and evaluation. Columns progress from full RGB (top left) to full thermal (bottom right) in 10\% increments.}
    \label{fig:fusion_levels}
\end{figure}

\subsection{Model training}
\label{sec:model_training}

Each fine-tuned YOLOv11 nano detector is trained on fused images corresponding to a single fusion level and a single illumination category, so that every model specializes in one operating regime. Two additional control models, described in Section~\ref{sec:experimental_variables},provide baselines for comparison. All fine-tuned models use the same training hyperparameters, summarized in Table~\ref{tab:training_hparams}.

\begin{table}[H]
\centering
\renewcommand{\arraystretch}{1.25}
\setlength{\tabcolsep}{10pt}
\caption{Training hyperparameters for all YOLOv11 nano detectors.}
\label{tab:training_hparams}
\begin{tabular}{cc}
\toprule
\textbf{Hyperparameter} & \textbf{Value} \\
\midrule
Optimizer      & AdamW \\
Learning rate  & 0.002 \\
Momentum       & 0.9 \\
Weight decay   & 0.0005 (weights only) \\
Image size     & $960 \times 960$ pixels \\
Batch size     & 16 \\
Maximum epochs & 30 \\
Early stopping & 5 epochs without validation improvement \\
\bottomrule
\end{tabular}
\end{table}

These settings are chosen to provide stable convergence, preserve object detail at the mug scale, and remain feasible within the available graphics processing unit (GPU) memory and computation budget.

\subsection{Evaluation framework}
\label{sec:experimental_protocol}

Evaluation proceeds in two stages. In the first stage, each fine-tuned model is evaluated on test sequences that match its illumination category, while the control models are evaluated across all categories. During dynamic trials, the turret tracks a moving mug for ten seconds. Frames with clearly spurious detections are going to be excluded so that occasional tracking glitches or false positives do not bias the mean confidence statistics.

In the second stage, we summarize performance using mean detection confidence per trial, defined as the average confidence across all detected frames in that trial. This metric reflected how strongly the model classified the target as a mug under a given configuration. Within each illumination category, models are ranked by their mean confidence, and the highest-ranked model in each category is selected as the candidate for adaptive runtime selection.

\subsection{Adaptive selection and runtime integration}
\label{sec:adaptive_framework}
At runtime, lux readings from the VEML7700 are mapped to an illumination category. The server loads the preselected top-performing model for the category and performs inference on incoming frames. Detection outputs are returned to the Pi, which translates the bounding box centers into pan and tilt commands for the stepper motors. Model switching occurs deterministically when lux readings cross category thresholds, ensuring reliable adaptation to lighting conditions.

\section{Results}
\label{sec:results}

To quantify the effect of RGB–LWIR fusion on object detection performance, we trained 33 YOLOv11n models, one for each of the 11 fusion levels ($\alpha = 1.0$ to $0.0$ in 0.1 increments) under three illumination categories: full-light, dim-light, and no-light. Each model’s performance was evaluated by computing the mean detection confidence across all six mug colors in the test set, a metric representing the model’s certainty in detecting a mug.

Table~\ref{tab:raw-confidence} summarizes the raw detection confidence for each cup color across all RGB--LWIR fusion levels and lighting conditions, including the YOLOv5n and YOLOv11n baselines. These values provide a complete numeric basis for the visualizations and aggregate metrics presented in the remainder of this section.

\begin{table}[H]
    \centering
    \caption{Detection confidence by cup color and RGB--LWIR fusion level across full-light, dim-light, and no-light conditions. Teal and yellow mugs, denoted by an asterisk ($^*$), were held out from training and used only for testing purposes.}
    \label{tab:raw-confidence}
    \includegraphics[width=\textwidth]{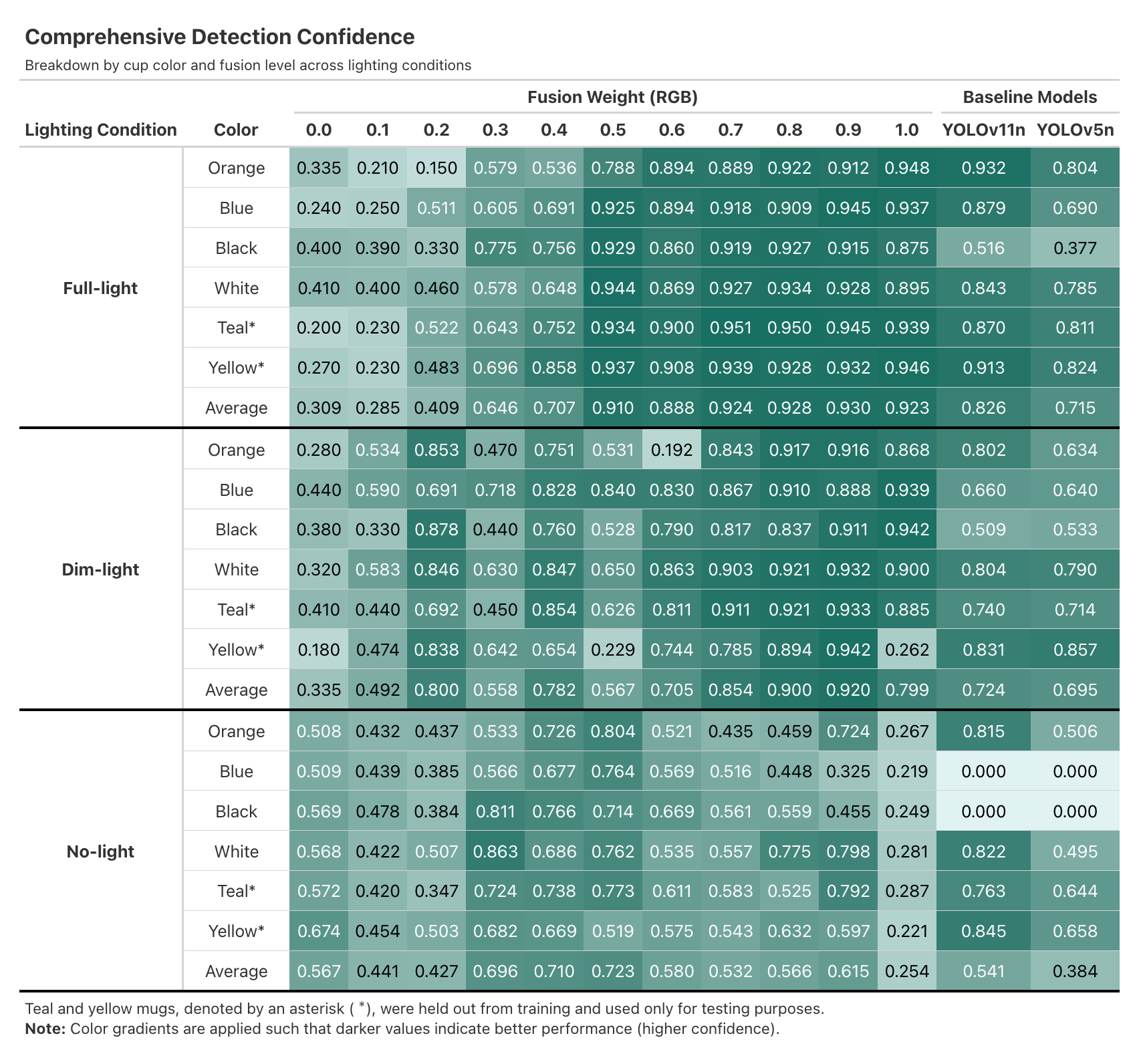}
\end{table}

\begin{figure}[H]
    \centering
    \includegraphics[width=\textwidth,keepaspectratio]{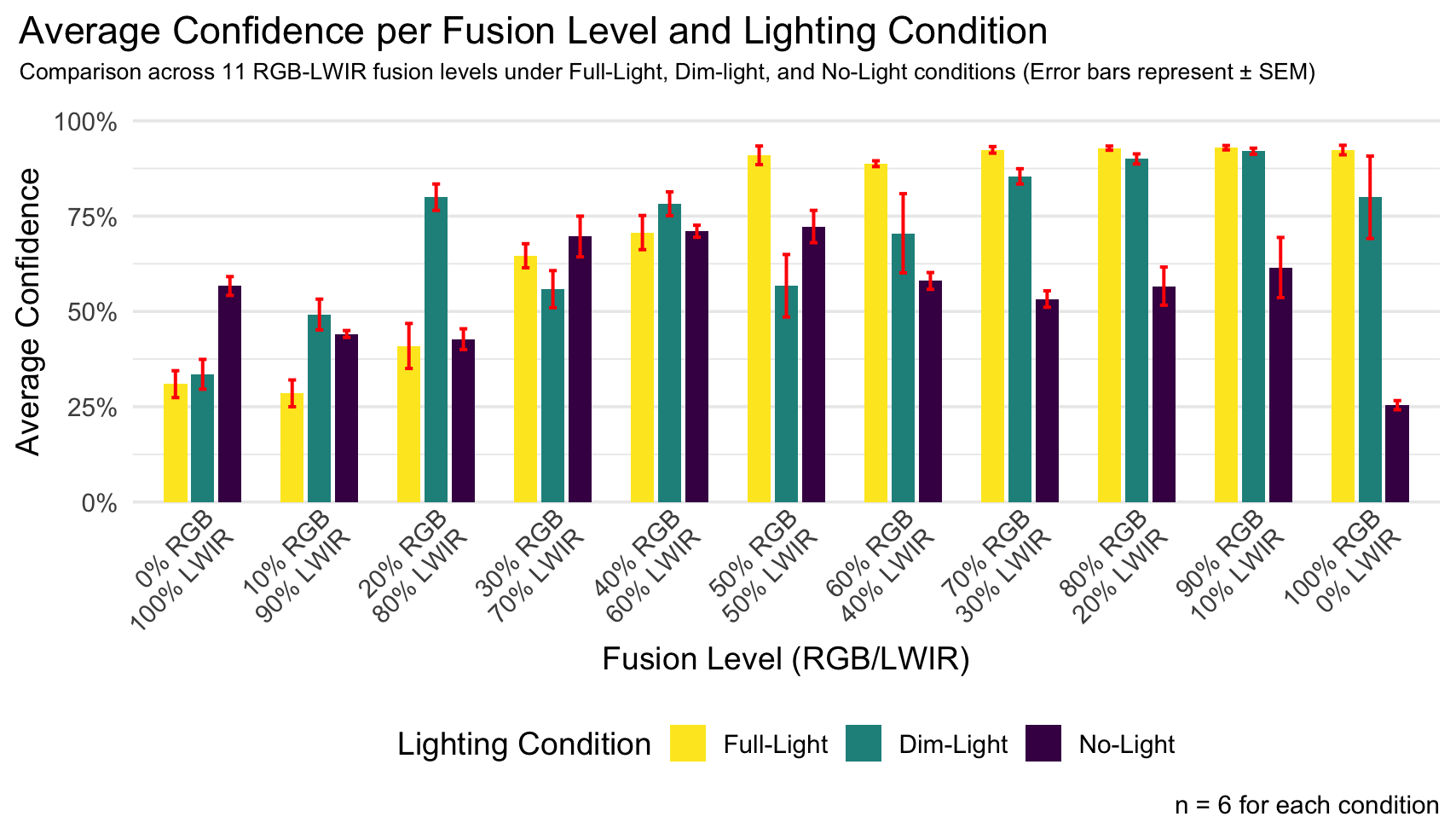}
    \caption{Average detection confidence of YOLOv11n models across eleven RGB–LWIR fusion levels under three lighting conditions. Higher values indicate greater model certainty in identifying mug targets. Error bars represent the standard error of the mean (SEM), as mean $\pm$ SEM, calculated as $\sigma / \sqrt{n}$.}
    \label{fig:avg_confidence}
\end{figure}

To visualize how detection confidence varies with fusion level and lighting overall, we next aggregated performance across mug colors. \autoref{fig:avg_confidence} plots these results, with each fusion level represented by a cluster of three bars: yellow (full-light), green (dim-light), and purple (no-light). The x-axis progresses from pure thermal (left) to pure RGB (right), and the y-axis shows the mean detection confidence. Error bars indicate the standard error of the mean (SEM), calculated as $\sigma / \sqrt{n}$.

Several clear patterns emerge.

Firstly, full-light (yellow bars) shows the strongest performance at or near pure RGB ($\alpha \approx 1.0$), with confidence gradually decreasing as thermal contribution increases. Specifically, the 80\% RGB / 20\% LWIR model achieved a mean confidence of 0.9283 ($\pm$0.0135), while 90\% RGB / 10\% LWIR reached 0.930 ($\pm$0.0142), only a +0.17\% relative increase. The 70\% RGB / 30\% LWIR model yielded 0.924 ($\pm$0.0212), representing a --0.45 percentage point drop compared with 80/20. Collectively, the 70--100\% RGB range all performed above 92\%, confirming that RGB-dominant fusion is highly effective under full illumination.

Secondly, no-light conditions (purple bars) exhibited a bell-shaped trend, with intermediate fusion levels performing best. Average confidence peaked at 0.723 ($\pm$0.104) for the 50\% RGB / 50\% LWIR model, closely followed by 0.710 ($\pm$0.0388) for the 40\% RGB / 60\% LWIR configuration. Although 50/50 achieved +0.0134 higher mean confidence (+1.72\% rel) than 40/60, the much lower variability of 40/60 resulted in the superior composite score (0.876 vs.\ 0.522). Confidence declined toward the extremes, with pure RGB (0.384) and pure LWIR (0.541) showing markedly lower values, demonstrating that both modalities alone are insufficient in complete darkness.

Finally, dim-light (green bars) generally followed the full-light trend, with RGB-dominant fusion performing best. The 90\% RGB / 10\% LWIR model achieved a mean confidence of 0.920 ($\pm$0.0200), the highest of all dim-light configurations, outperforming the 80/20 ratio (0.9000 $\pm$0.0333) by +0.0203 (+2.26\% rel) and the 70/30 ratio (0.854 $\pm$0.0491) by +0.0660 (+7.73\% rel). This indicates that a modest thermal contribution can enhance performance, but excessive thermal weighting reduces detection reliability in low-light scenarios.

\autoref{fig:avg_confidence} illustrates that fusion performance is strongly illumination-dependent: full-light and dim-light trends are left-skewed, favoring RGB-heavy configurations, whereas no-light conditions benefit most from intermediate blending. These observations highlight the effectiveness of selectively adjusting the RGB–LWIR ratio according to lighting, supporting the adaptive fusion strategy implemented in the turret system.

\begin{figure}[H]
    \centering
    \includegraphics[width=\textwidth,keepaspectratio]{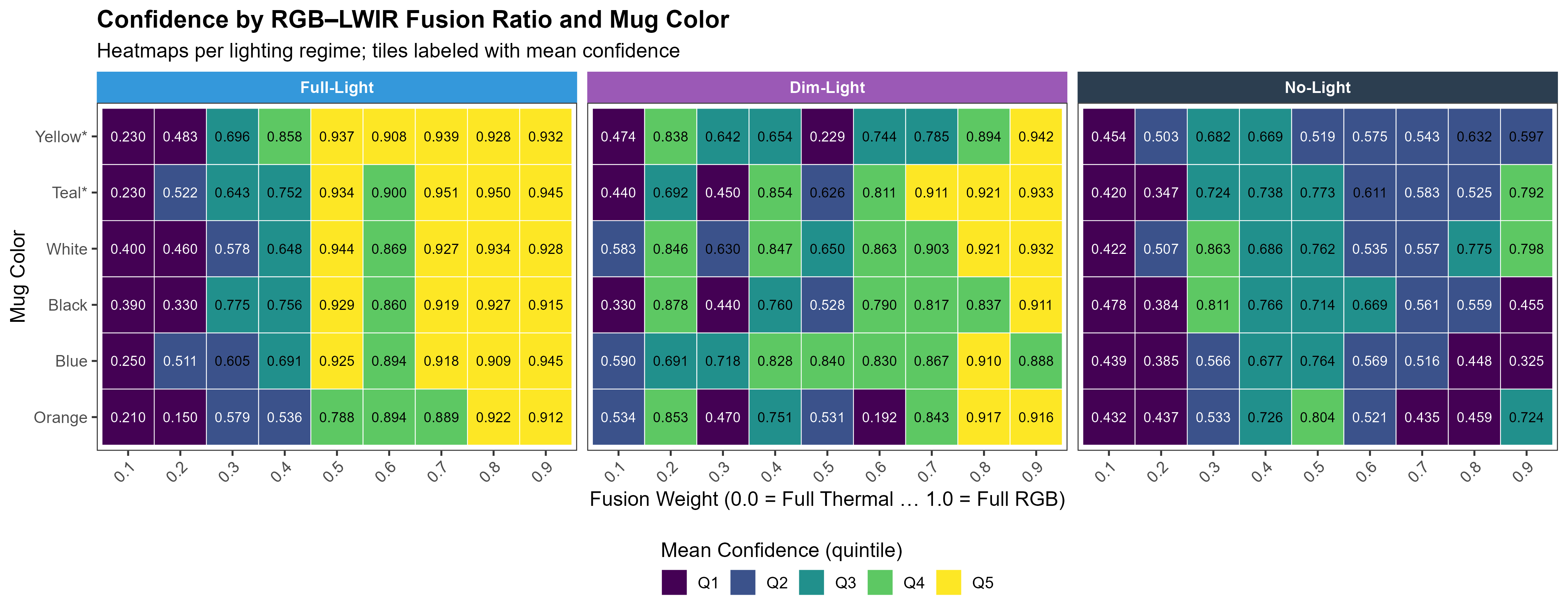}
    \caption{Heatmap visualization of mean detection confidence across RGB--LWIR fusion ratios and mug colors, separated by lighting regime. 
    Each panel corresponds to one illumination condition (full-light, dim-light, no-light), with color indicating mean confidence and numeric labels showing exact values. 
    The quintile-based color scale (Q1--Q5) highlights relative performance tiers, enabling direct comparison across panels.}
    \label{fig:fusion_heatmaps}
\end{figure}

To understand how these global trends vary by mug color, we next examine color- and fusion-specific mean confidence values. \autoref{fig:fusion_heatmaps} summarizes these values as heatmaps, with one panel per illumination regime and cells corresponding to a particular color--fusion combination. This representation highlights relative performance tiers and makes it easy to compare which colors benefit most from RGB-heavy versus thermally weighted fusion under each lighting condition.

In dim-light, the optimal fusion region remains RGB-heavy but shifts slightly toward intermediate ratios. Peak confidence for most colors occurs between 0.7 and 0.9 fusion weight, though performance at 0.5-0.6 remains competitive for several colors. Compared to full-light, confidence values show greater dispersion across fusion ratios, reflecting increased sensitivity to modality balance as RGB quality degrades. Notably, darker mugs (blue and black) exhibit larger performance drops at low fusion weights, whereas lighter or more saturated colors (white, yellow, teal) retain high confidence across a broader range of RGB-dominant settings.

In no-light, the structure of the heatmap changes more significantly. Peak confidence no longer occurs at RGB-dominant fusion; instead, intermediate fusion ratios (approximately 0.3--0.5 RGB) consistently yield the strongest performance across most mug colors. Pure or near-pure RGB fusion (fusion $\geq$ 0.7) shows a significant decline in confidence, confirming that visible-spectrum information alone is insufficient in full darkness. At the same time, pure thermal fusion (fusion $\leq$ 0.2) also underperforms, suggesting that while LWIR provides essential saliency, limited RGB contribution still supplies structural context that stabilizes detections.

Across all illumination regimes, mug color affects detection confidence and how sensitive performance is to the fusion ratio. Lighter colors such as white and yellow mugs consistently occupy the highest-confidence tiers, while darker colors such as black and blue mugs remain the most challenging, particularly in dim and no-light conditions. Importantly, the held-out teal and yellow mugs follow the same fusion-dependent patterns as the trained colors, indicating that the learned fusion behavior generalizes beyond the training set rather than overfitting to specific appearances. Overall, the heatmap provides strong visual evidence that no single fusion ratio is optimal across lighting conditions, directly motivating the adaptive fusion strategy used in the turret system.

\begin{figure}[H]   
    \centering
    \includegraphics[width=\textwidth,keepaspectratio]{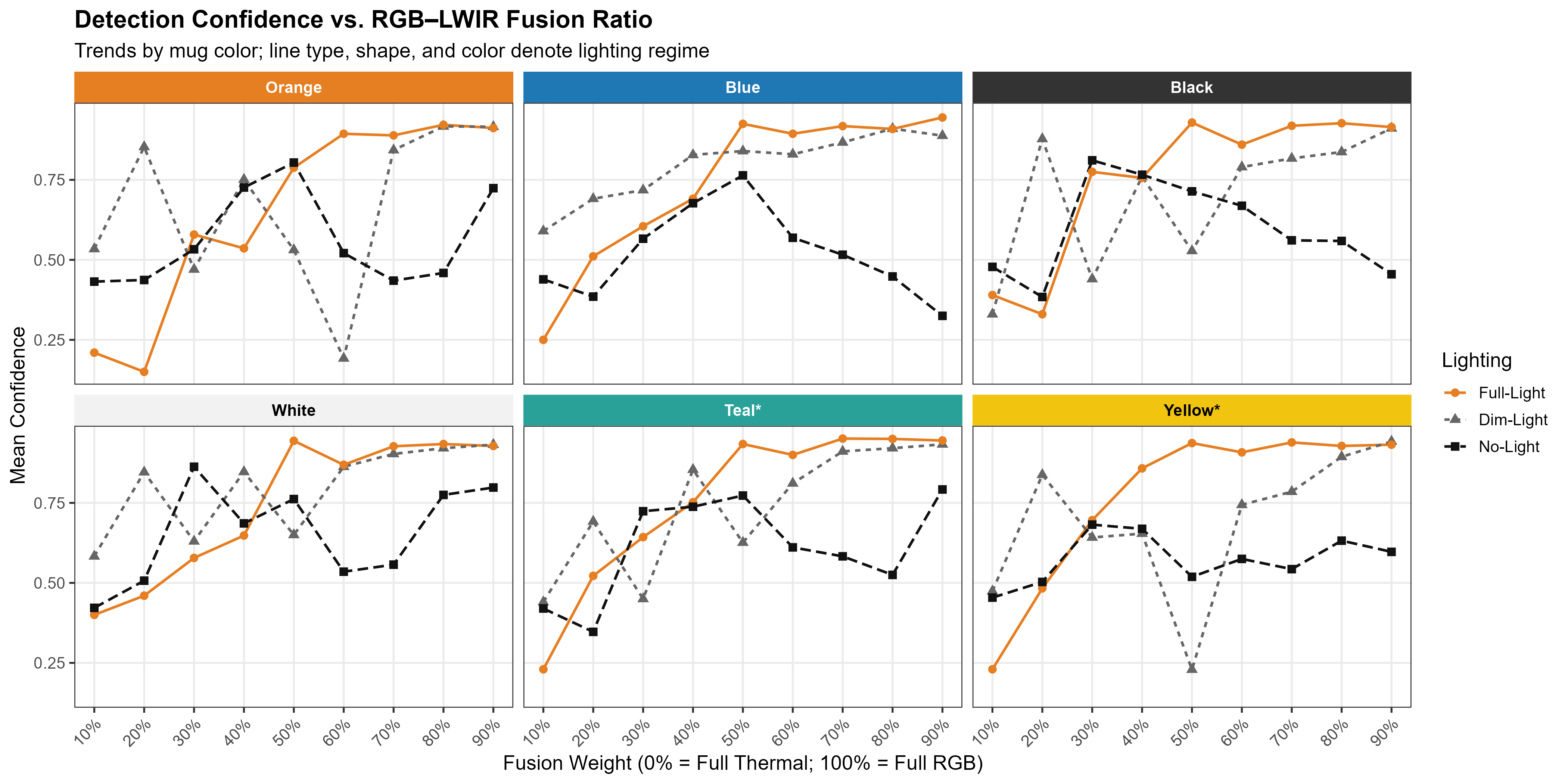}
    \caption{Line chart showing mean detection confidence across RGB--LWIR fusion ratios for each mug color, with separate panels and color-coded headers. 
    The x-axis represents the fusion weight (0.0 = full thermal, 1.0 = full RGB), and the y-axis represents mean confidence. 
    Line type and marker indicate lighting conditions (full-light, dim-light, no-light), highlighting performance trends across illumination regimes.}
    \label{fig:mugs_trend_smallmultiples}
\end{figure}

To better visualize how each mug color responds to changes in the fusion ratio across lighting regimes, \autoref{fig:mugs_trend_smallmultiples} plots mean confidence as a function of RGB weight for each color, with separate panels and line styles for the three illumination conditions. This view emphasizes the trajectories of individual colors rather than their relative tiers.

Figure~\ref{fig:mugs_trend_smallmultiples} shows how each mug color responds to the RGB--LWIR fusion weight under the three lighting regimes. In full-light, all colors follow a clear upward trend as the fusion becomes more RGB-heavy: confidence is relatively low at 10--20\% RGB, rises quickly by 30--50\% RGB, and then largely plateaus at high confidence from 50--90\% RGB. In dim-light, the same general pattern holds (better performance at higher RGB weights), but the curves are more irregular, indicating that small changes in fusion ratio can affect stability when RGB quality begins to degrade. 

In no-light, the behavior shifts: for several colors (especially blue, black, and yellow), confidence is highest at \emph{intermediate} fusion weights (roughly 30--50\% RGB) and drops as the input becomes heavily RGB-dominant (70--90\% RGB). This supports the key takeaway that the best fusion ratio depends on illumination: RGB-dominant fusion works best when light is available, while mixed fusion is more reliable in darkness. The held-out colors (teal* and yellow*) follow the same regime-dependent shapes, suggesting the fusion trends generalize beyond the training colors rather than being tied to a specific appearance.

To provide a summary of detection performance across all cup colors, Table~\ref{tab:color-means} reports the overall mean detection confidence values, averaged across all 39 model-condition combinations (33 fine-tuned YOLO models and 2 COCO YOLO models, each tested in 3 lighting conditions).

\begin{table}[H]
  \centering
  \caption{Average Detection Confidence by Cup Color (Averaged Across All Models and Conditions)}
  \label{tab:color-means}
  \begin{tabular}{lc}
    \toprule
    \textbf{Cup Color} & \textbf{Mean Confidence (\%)} \\
    \midrule
    \multicolumn{1}{c}{White}    & 70.98 \\
    \multicolumn{1}{c}{Teal}     & 68.75 \\
    \multicolumn{1}{c}{Yellow}   & 66.07 \\
    \multicolumn{1}{c}{Orange}   & 63.22 \\
    \multicolumn{1}{c}{Blue}     & 63.21 \\
    \multicolumn{1}{c}{Black}    & 61.13 \\
    \bottomrule
  \end{tabular}
\end{table}

As indicated in Table~\ref{tab:color-means}, white cups were the only color to achieve a mean confidence above 70\%, making them the most reliably detected. Teal and yellow followed, at 68.75\% and 66.07\%, respectively, representing relative decreases of 3\% and 7\% compared with white. Orange and blue cups showed very similar performance, with only a 0.01\% difference. Black cups were the lowest-performing color, at 61.13\%, roughly 10\% below the top-performing white cups.

These results demonstrate that lighter and more saturated colors (white, teal, yellow) consistently yield higher detection confidence, while darker colors (blue, black) are more challenging for the models, particularly under dim or no lighting.

To compare model performance across these varying conditions in a more holistic way, we introduced a composite score. This single measure accounts for both detection certainty and consistency across mug colors by combining each model's mean confidence and standard deviation using the composite score defined in Equation~\ref{eq:composite-score}. This approach allows us to identify models that are not only accurate but also stable in performance.

\medskip

\begin{equation}
\label{eq:composite-score}
\Large
\frac{\bar{\mu} - \bar{\mu}_{\min}}{\bar{\mu}_{\max} - \bar{\mu}_{\min}}
\;-\;
\frac{\sigma - \sigma_{\min}}{\sigma_{\max} - \sigma_{\min}}
\end{equation}

\medskip

In this equation, $\bar{\mu}$ denotes the mean detection confidence for a model, and $\sigma$ is the standard deviation of that confidence across trials. Here, $\bar{\mu}_{\min}$ and $\bar{\mu}_{\max}$ represent the lowest and highest mean detection confidence observed among all models for the given lighting condition, respectively. Similarly, $\sigma_{\min}$ and $\sigma_{\max}$ denote the minimum and maximum standard deviation across all models under the same lighting condition. Each term is normalized by subtracting the minimum observed value and dividing by the range across all models. The first term rewards models with higher confidence, while the second term penalizes models with greater variability. Subtracting the normalized variability from the normalized confidence yields a composite score that favors models that are both accurate and consistent.

Using this composite score, we ranked all 33 fusion models under each lighting condition. Table~\ref{tab:top3-fusion-levels} presents the top three fusion configurations for full-light, dim-light, and no-light conditions. In the table, the model marked with an asterisk (*) denotes the highest composite score within each lighting condition.

\begin{table}[H]
    \centering
    \caption{Top three fusion levels by illumination condition.}
    \label{tab:top3-fusion-levels}
    \includegraphics[width=\textwidth]{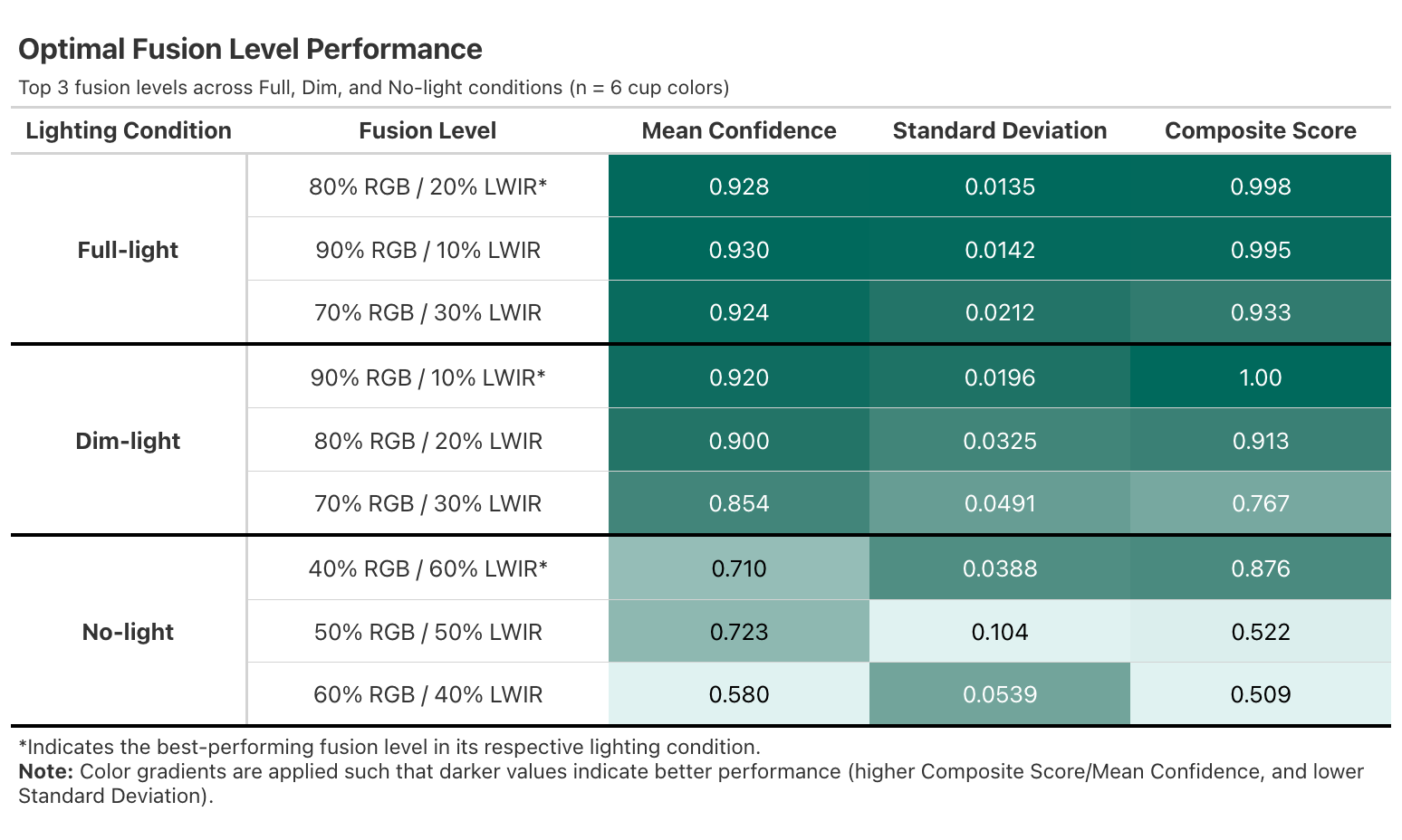}
\end{table}

Under full-light conditions, the best-performing model was 80\% RGB / 20\% LWIR, which outperformed the second-ranked 90\% RGB / 10\% LWIR by a margin of 0.0031 in composite score. Although the second-ranked model achieved a slightly higher average confidence, the first-ranked model exhibited a lower standard deviation, resulting in a superior overall composite score.

Under dim-light conditions, the top model was 90\% RGB / 10\% LWIR, which achieved the maximum composite score of 1.0. This is because it had the highest average confidence and lowest variability out of all the dim-light models.

Under no-light conditions, the best-performing model shifted significantly toward LWIR, with 40\% RGB / 60\% LWIR taking the lead. Its composite score was much higher than the second-ranked and third-ranked models, which had scores of 0.5219 and 0.5093, respectively.

\begin{figure}[H]
    \centering
    \includegraphics[width=\textwidth,keepaspectratio]{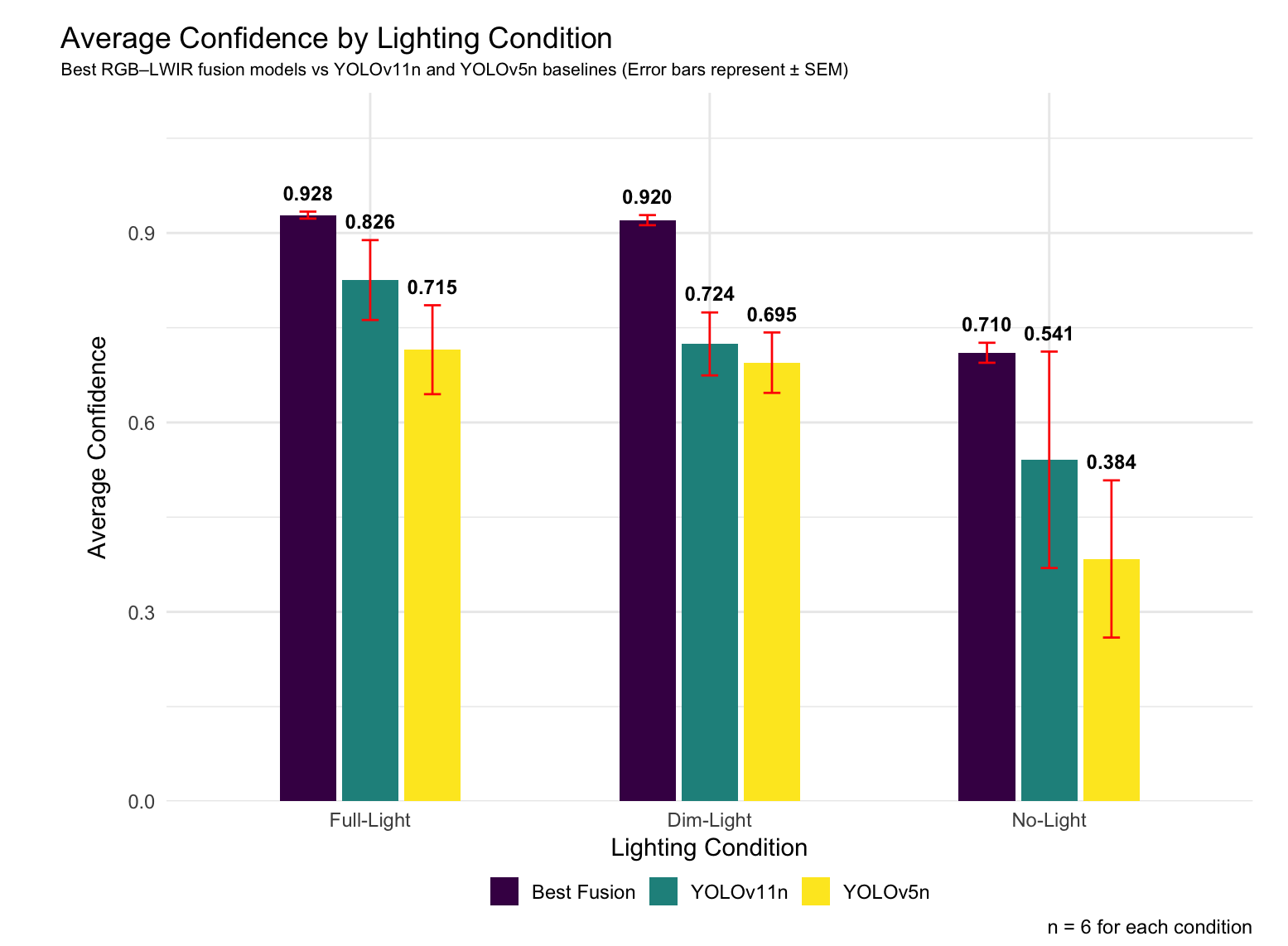}
    \caption{Average confidence across lighting conditions for the best RGB--LWIR fusion model in each lighting regime and two YOLO baselines (YOLOv11n and YOLOv5n). Bars show mean confidence, and error bars indicate $\pm$ standard error of the mean (SEM) computed from six trials per condition.}
    \label{fig:overall_confidence}
\end{figure}

Building on these results, we next compared the best-performing fusion models identified in Table~\ref{tab:top3-fusion-levels} against the baseline YOLO models (YOLOv5n and YOLOv11n). This comparison allows us to assess whether the gains achieved through fusion hold up when compared to leading single-modality detectors across different lighting conditions. Figure~\ref{fig:overall_confidence} summarizes these comparisons, showing mean confidence by model and lighting condition with $\pm$ standard error of the mean (SEM) error bars (six trials per condition). Numeric labels on each bar report the mean confidence values.

In both full-light and dim-light conditions, the best fusion models achieved the highest average confidence, reaching 0.928 and 0.920 respectively, compared to 0.826 and 0.724 for YOLOv11n, and 0.715 and 0.695 for YOLOv5n. The non-overlapping SEM error bars in Figure~\ref{fig:overall_confidence} indicate these improvements are statistically significant. In no-light conditions, the best fusion model maintained a confidence of 0.710, outperforming YOLOv11n (0.541) and YOLOv5n (0.384), although partial overlap of error bars suggests a smaller performance margin. Across all lighting conditions, fusion consistently outperformed single-modality baselines, with the largest relative gains observed under low illumination.

\section{Discussion}
\label{sec:discussion}
The results demonstrate that adaptive RGB--LWIR fusion offers clear and consistent advantages over single-modality detection across illumination conditions, consistent with prior aerial and ground-based multispectral studies where fused RGB--thermal inputs outperform either modality alone across day and night \cite{ref2,nataprawira2021multispectral}. In full-light, fusion improves mean confidence by +10.2~percentage points (pp) over YOLOv11n and +21.3~pp over YOLOv5n; in dim-light, by +19.7~pp and +19.6~pp; and in no-light, by +16.9~pp and +32.6~pp (Fig.~\ref{fig:overall_confidence}). These gains align with fusion-level trends in Fig.~\ref{fig:avg_confidence}: RGB-heavy ratios dominate when light is available, whereas intermediate ratios (40--60\% RGB) recover performance in darkness, mirroring observations that multispectral YOLO variants typically close the no-light performance gap relative to RGB-only baselines through attention-guided and illumination-aware fusion \cite{ref11,cao2021attention}. Across all lighting conditions, fusion-based YOLOv11n models achieved higher mean detection confidence than both YOLOv5n and YOLOv11n baselines trained exclusively on RGB images. The improvement was most pronounced in low-light scenarios, where thermal imagery contributed critical features absent in RGB input, similar to Gallagher and Oughton’s finding that fused RGB--LWIR detections substantially outperform RGB in pre-sunrise and post-sunset flights \cite{ref2}. These results, together with the broader multispectral survey evidence that fixed fusion strategies rarely dominate across conditions \cite{ref1}, suggest that a single, fixed fusion ratio is suboptimal; instead, dynamically selecting the ratio and corresponding model can maximize robustness across environments.

The optimal fusion ratio changes with lighting because RGB and LWIR each contribute different types of information depending on illumination. In full-light, RGB provides most of the useful features for detection such as clear color, fine texture, and sharp edges, while a small amount of LWIR can help highlight heat-emitting surfaces, consistent with reports that RGB and fusion track closely in full-light while thermal primarily adds value for high-contrast heat targets \cite{ref2,ref16}. Empirically, RGB-dominant fusion (70--100\% RGB) sustained $\geq$92\% mean confidence, with the best full-light model at 80/20 reaching 0.9283 (vs.\ 0.9295 at 90/10, a +0.17\% relative uptick), and 70/30 at 0.9238 (--0.45~pp vs.\ 80/20). In dim-light, the quality of RGB data decreases due to sensor noise, so a slightly higher LWIR proportion improves object boundaries and separation from the background, in line with illumination-aware fusion systems that shift weight toward thermal cues at night to reduce miss rate \cite{ref9,ref11}. Consistent with this, the 90/10 model achieved 0.9203 ($\pm$0.0196~SEM), outperforming 80/20 (0.9000, $\pm$0.0325) by +0.0203 (+2.26\% rel) and 70/30 (0.8543, $\pm$0.0491) by +0.0660 (+7.73\% rel). In no-light conditions, RGB offers minimal usable detail, making LWIR the main source for detection, with a small RGB contribution still helpful for structural context when thermal images are affected by reflections or background clutter \cite{ref2,ref18}. Here, intermediate blends were superior: 40/60 produced 0.7103 ($\pm$0.0388) and 50/50 reached 0.7227 ($\pm$0.1039), both well above the pure modalities (RGB 0.384; LWIR 0.541). Notably, despite 50/50’s slightly higher mean, 40/60’s much lower variability yielded a substantially higher composite score (0.876 vs.\ 0.522), reinforcing the preference for intermediate fusion in darker conditions.

During early experiments, we initially trained 11 fusion models, each corresponding to a specific RGB--LWIR ratio but pooled across all illumination conditions. However, these pooled models struggled to develop a strong understanding of the training data, as the large variation across full-light, dim-light, and no-light settings made it difficult for a single model to capture consistent patterns. This is consistent with multispectral survey findings that domain shift across weather, lighting, and collection setups can undermine performance when a single detector is expected to generalize across all regimes \cite{ref4,kim2022uncertainty}. To address this, we adopted a revised strategy and trained 33 models, each tied to both a specific fusion ratio and a single illumination condition. This approach substantially improved performance consistency, as models were able to learn condition-specific features without interference from other environments.

Unlike many existing RGB--LWIR fusion approaches, which use a single dual-stream model trained across all lighting conditions with largely fixed fusion operators \cite{ref23,ref25} or rely on feature-level attention mechanisms to dynamically weight modalities within a shared backbone \cite{ref11,ref27}, our method adopts a lighting-specific model specialization strategy. By training 33 dedicated YOLOv11n models, one for each fusion ratio within each illumination category, and using a lux sensor to determine light intensity, our system can switch models in real time without manual intervention. Recent YOLO-based multispectral detectors similarly emphasize architecture choices that balance fusion quality with deployment latency on embedded platforms, for example by compressing backbones for low-latency multispectral driving and aerial detection \cite{roszyk2022yolov4,shao2022airyolov3}. Our design contrasts with more complex attention- or transformer-based fusion \cite{ref11,ref27} and aligns with Gallagher and Oughton’s call for adaptive architectures and deployment-aware designs in multispectral YOLO systems \cite{ref1}. It also simplifies deployment on embedded systems, as model switching is based on a single sensor reading rather than online feature reweighting, and our results show that it is responsive to real-world lighting variability.

Because the composite score prioritizes both accuracy (mean) and standard deviation, it directly supports deployment choices: the system can prefer 80/20 in full-light, 90/10 in dim-light, and 40/60 in no-light to maximize confidence while minimizing variability across target appearances.

Despite the improved performance achieved through adaptive model selection, several limitations were observed that may affect the system’s generalizability. First, the dataset used for training primarily consisted of static images of the target objects in fixed positions, whereas the testing phase involved dynamic motion across the frame. This train--test domain shift reduced detection confidence when targets moved rapidly, appeared near the frame edges, or were partially occluded, echoing broader concerns that curated benchmarks often fail to capture the variability and occlusion patterns seen in real deployments \cite{ref4,kim2022uncertainty}. Second, manual annotation errors, such as including portions of the black rotating platform in bounding boxes, introduced label noise, potentially biasing the models toward non-target features; similar labeling and alignment challenges have motivated automated cross-modal transfer pipelines such as MATT and other sensor-fusion reviews \cite{ref3,ref8}. While the fusion trends were consistent, some per-color confidence curves showed noticeable variability across fusion ratios. This is likely a result of limited sample sizes within each color setting. In addition, LWIR imagery in this study was represented using a single fixed color palette from the thermal camera output. Different thermal palettes can alter apparent color and contrast in LWIR frames, which may influence learned features in fused RGB--LWIR inputs and interact with object color. Evaluating different thermal palettes is, therefore, an important direction for future work. Collecting more data per condition would further help clarify these trends.
Finally, LWIR imagery was occasionally affected by sensor noise and thermal reflections, such as glints from hot water surfaces, which sometimes produced false positives, consistent with reported issues of thermal auto-gain, non-uniformity correction, and parallax in RGB--thermal systems \cite{ref2,ref8}. Addressing these limitations will be important for improving accuracy.

\section{Conclusion}
\label{sec:conclusion}

The proposed illumination-aware RGB--LWIR fusion framework demonstrated consistent improvements in detection performance over single-modality baselines across all lighting conditions, with gains of up to 30\% in mean detection confidence against base YOLO models. By dynamically selecting among the best fusion ratios (80/20 for full-light, 90/10 for dim, 40/60 for no-light), we achieved up to 30\% absolute gains in mean confidence over baseline YOLO models, with statistically significant improvements (non-overlapping SEM) under full-light and dim-light. Optimal fusion ratios varied with illumination, 80\% RGB / 20\% LWIR in full-light, 90\% RGB / 10\% LWIR in dim-light, and 40\% RGB / 60\% LWIR in no-light, highlighting the advantage of switching between lighting-specific models. By using a low-cost lux sensor with a subset of pre-trained fusion models, the system achieved real-time adaptation without manual intervention or computationally expensive feature reweighting. Additionally, the model showed high confidence even on object colors absent from the training data. These capabilities are directly relevant to mission-critical applications such as autonomous security turrets, mobile robotics, and search-and-rescue operations, where lighting can shift abruptly.

Future work will focus on enhancing both the diversity of the training dataset and the temporal reasoning capabilities of the detection framework. Expanding the dataset to include targets undergoing varied motions, partial occlusions, and interactions with cluttered backgrounds will help mitigate the train--test domain gap identified in this study. Data augmentation techniques such as motion blur, contrast jittering, and geometric masking will be applied to simulate real-world imaging conditions more effectively. In addition, transitioning from a frame-by-frame detection paradigm to video-based processing using temporal architectures, such as Convolutional Long Short-Term Memory (ConvLSTM) networks or 3D Convolutional Neural Networks (3D CNNs), could improve stability by leveraging motion continuity across frames . Finally, integrating false positive regions identified during testing back into the training loop will enable the models to explicitly learn to suppress high-confidence misdetections, thereby refining bounding box precision.

\section*{Acknowledgment}
This research was made possible through the support of George Mason University’s College of Science, which
supports the ASSIP Program. We would like to thank the Geography and Geoinformation Science Department
at George Mason University for funding this research. Additional thanks to Dr. Edward Oughton, Mr. James
Gallagher, and Ms. Tyler Treat whose guidance contributed to the success of this project.

\section*{Data and Code Availability}

\noindent
The dataset used in this study is publicly available on Zenodo at:\\
\url{https://doi.org/10.5281/zenodo.18020084}

\medskip
\noindent
The associated code for data collection, fusion, and evaluation is available at:\\
\url{https://github.com/IRIS2Lab/Team-3}

\end{document}